\documentclass{article}

\usepackage{arxiv}

\usepackage[utf8]{inputenc} 
\usepackage[T1]{fontenc}    
\usepackage{hyperref}       
\usepackage{url}            
\usepackage{booktabs}       
\usepackage{amsfonts}       
\usepackage{nicefrac}       
\usepackage{microtype}      
\usepackage{xcolor}         

\usepackage{orcidlink}

\usepackage[textsize=tiny]{todonotes}

\usepackage{booktabs}  
\usepackage{algorithm}
\usepackage{algpseudocode}
\usepackage{subcaption}
\usepackage{caption}
\usepackage{bm}
\usepackage{natbib}
\usepackage{wrapfig}
\setcitestyle{numbers,square}

\renewcommand{\gg}{\mathbf{g}}

\usepackage{multirow}
\usepackage{enumitem}
\usepackage{makecell}
\usepackage{tabularx} 
\usepackage{arydshln}
\usepackage{pifont}
\usepackage{soul}
\usepackage{tikz}
\usepackage{amsmath}
\usetikzlibrary{decorations.pathreplacing,calc}
\newcommand{\tikzmark}[1]{\tikz[overlay,remember picture] \node (#1) {};}
\newcommand*{\AddNote}[4]{%
	\begin{tikzpicture}[overlay, remember picture]
		\draw [decoration={brace,amplitude=0.5em,raise=1ex},decorate,thick,gray]
		($(#3)!(#2.south)!($(#3)-(0,1)$)$) --
		($(#3)!(#1.north)!($(#3)-(0,1)$)$)
		node [align=center, text width=3cm, pos=0.5, anchor=east, font=\it, rotate=90, yshift=4ex, xshift=11ex] {#4};
	\end{tikzpicture}
}%

\newcommand{\xmark}{\ding{55}}
\newcommand{\cmark}{\ding{51}}

\newcommand{\algname}{Decay Pruning Method{}}
\newcommand{\algacro}{DPM{}}

\title{\algname:  Smooth Pruning with a Self-Rectifying Procedure}

\date{}

\newif\ifuniqueAffiliation
\uniqueAffiliationtrue

\ifuniqueAffiliation 
\author{ Minghao Yang \\
	Ningbo University\\
	\texttt{216003669@nbu.edu.cn} \\
	\and
	\textbf{ Linlin Gao}\\
	 Ningbo University\\
	 	\texttt{gaolinlin@nbu.edu.cn} \\
	 	\AND
	 	\textbf{Pengyuan Li}\\
	 	IBM Research\\
	 	\texttt{pengyuandan@ibm.com} \\
	 	 \and
	 	\textbf{Wenbo Li}\\
	 		Zhejiang Lab\\
	 	\texttt{liwenbo\_923@hotmail.com} \\
	 	 \and
	 	\textbf{Yihong Dong}\\
	 	 Ningbo University\\
	 	\texttt{dongyihong@nbu.edu.cn} \\
	 	 \AND
	 	\textbf{Zhiying Cui}\\
	 	Ningbo University\\
	 	\texttt{226004124@nbu.edu.cn} \\
	}

\else
\usepackage{authblk}

\newbox{\orcid}\sbox{\orcid}{\includegraphics[scale=0.06]{orcid.pdf}} 
\author[1]{%
	\href{https://orcid.org/0000-0000-0000-0000}{\usebox{\orcid}\hspace{1mm}David S.~Hippocampus\thanks{\texttt{hippo@cs.cranberry-lemon.edu}}}%
}
\author[1,2]{%
	\href{https://orcid.org/0000-0000-0000-0000}{\usebox{\orcid}\hspace{1mm}Elias D.~Striatum\thanks{\texttt{stariate@ee.mount-sheikh.edu}}}%
}
\affil[1]{Department of Computer Science, Cranberry-Lemon University, Pittsburgh, PA 15213}
\affil[2]{Department of Electrical Engineering, Mount-Sheikh University, Santa Narimana, Levand}
\fi

\begin{document}
	\maketitle

\begin{abstract}
	
	Current  structured pruning methods often result in considerable accuracy drops due to abrupt network changes and loss of information from pruned structures.  To address these issues, we introduce the \algname{}  (\algacro), a novel smooth pruning approach with a self-rectifying mechanism. \algacro{}  consists of two key components: \textit{(i) Smooth Pruning:} It converts conventional single-step pruning into  multi-step smooth pruning, gradually reducing redundant structures to zero over $N$ steps with ongoing optimization.  \textit{(ii) Self-Rectifying:} This procedure  further enhances the aforementioned process  by rectifying sub-optimal pruning decisions based on gradient information. 	Our approach demonstrates strong generalizability and can be easily integrated with various existing pruning methods. We validate the effectiveness of \algacro{} by integrating it with three  popular pruning methods: OTOv2, Depgraph, and Gate Decorator. Experimental results show consistent improvements in performance compared to the original pruning methods, along with further reductions of FLOPs in most scenarios. Our codes are available at \href{https://github.com/Miocio-nora/Decay\_Pruning\_Method.git}{https://github.com/Miocio-nora/Decay\_Pruning\_Method}.
\end{abstract}

\section{Introduction}\label{sec:intro}
Deep Neural Networks (DNNs) have been widely used for various applications, such as image classification \cite{li2023yolov6,wang2023tracking}, object segmentation \cite{ronneberger2015unet,sofiiuk2021reviving}, and object detection \cite{Cheng2024YOLOWorld,zhao2024open}. However, the increasing size and complexity of DNNs often result in substantial computational and memory requirements, posing challenges for deployment on resource-constrained platforms, such as mobile or embedded devices. Consequently, developing efficient  methods to reduce the computational complexity and storage demands of large models, while minimizing performance degradation, has become essential.

Network pruning is one of the most popular methods in model compression. Specifically, current network pruning methods are categorized into unstructured and structured pruning \cite{Cheng2023ASO}.  Unstructured pruning \cite{frantar2022spdy,AdaPrune2021}   focuses on eliminating individual weights from a network to create fine-grained sparsity. Although these approaches  achieve an excellent balance between model size reduction and accuracy retention, they often require specific hardware support for acceleration, which is impractical for general-purpose computing environments. Conversely, structured pruning \cite{li2017pruning,hu2016network,luo2017entropybased} avoids these hardware dependencies by eliminating redundant network structures,  thus introducing a more manageable and hardware-compatible form of sparsity. As a result, structured pruning has become popular and is extensively utilized.

\begin{figure*}[t]
	\centering
	\includegraphics[width=0.93\textwidth]{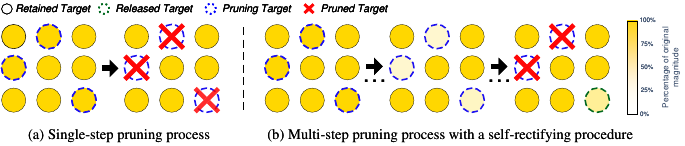}
	\begin{minipage}[b]{.5\linewidth}
		\phantomsubcaption 
		\label{fig:single-step}
	\end{minipage}%
	\begin{minipage}[b]{.5\linewidth}
		\phantomsubcaption 
		\label{fig:multi-step}
	\end{minipage}
	\caption{\textbf{Single-step pruning vs. Decay pruning method}. (a) Once pruning targets are identified, the single-step pruning process removes them in one action, risking abrupt network changes and irreversible information loss. (b) In contrast, the Decay Pruning Method gradually reduces the weight values of pruning structures over \(N\) steps, employing a gradient-driven self-rectifying procedure to identify and rectify sub-optimal decisions.}
	\label{fig:pruning-methods}

\end{figure*}

Existing structured pruning methods fall into two categories based on the granularity of the pruning targets: layer-wise and channel-wise pruning  \cite{Cheng2023ASO}.
Layer-wise pruning \cite{layerwise1,layerwise2} removes entire layers from a network, which significantly impacts the model performance due to its large-grained nature  \cite{zhu2017prune}.  On the other hand,  channel-wise pruning \cite{otov1,fang2023depgraph} offers a more  fine-grained strategy by removing channels.  This method brings less drastic network changes, achieving excellent trade-off between accuracy and model efficiency. Consequently, this paper focuses on channel-wise pruning.

However, current channel-wise pruning methods often apply a single-step pruning process, where redundant structures, once identified, are immediately removed or set to zero in a single operation. Figure \ref{fig:single-step} shows an illustration of typical single-step pruning process.  This approach often encounters several critical challenges: 

(1) \textbf{Accuracy damage from abrupt pruning}: 
Single-step pruning removes redundant network structures drastically, leading to unavoidable information loss and abrupt network changes. These abrupt losses and changes often result in a reduction in model performance, which is particularly hard to recover when pruning large-scale structures, even with subsequent fine-tuning \cite{zhu2017prune}. 

(2) \textbf{Sub-optimal  identification of redundant structures}: Furthermore, the single-step  process typically bases its pruning decisions on the current state or gradient information from a single batch \cite{fang2023depgraph,otov2}. Such an approach overlooks networks' future evolution and is susceptible to noise or peculiarities in the current data, leading to sub-optimal pruning decisions. 

Soft Filter Pruning (SFP)  \cite{He2018SoftFP} and various dynamic pruning strategies \cite{lin2020dynamic,evci2021rigging} have been developed to mitigate these issues by allowing for the optimization of pruned structures  for better information retaining, or enabling the regrowth of mistakenly pruned structures  for improved decision-making.  However, their persistent reliance on the single-step pruning process limits their effectiveness in fully aaddressing these challenges.

In response, we propose \algname{} (\algacro{}), a novel multi-step smooth pruning approach with a self-rectifying procedure. A high-level illustration of our DPM is shown in Figure \ref{fig:multi-step}. Specifically, \algacro{}  includes two main components: \textit{(i) \textbf{Smooth Pruning (SP)}}:  This procedure gradually decays the weights of redundant structures to zero over $N$ steps while maintaining continuous optimization.
This gradual approach avoids abrupt changes to the network and improves information retention during pruning. \textit{(ii) \textbf{Self-Rectifying (SR)}}: This procedure utilizes gradient information of the decaying structures to dynamically identify and rectify sub-optimal pruning decisions.  This approach takes into account the network's future evolving state, ensuring more reliable and adaptive pruning decisions. Our DPM can be seamlessly integrated into various existing pruning frameworks, resulting in significant accuracy improvements and further reductions  in FLOPs compared to original pruning methods. We evaluate the effectiveness and generalizability of \algacro{} across multiple pruning frameworks, including two auto pruning frameworks, OTOv2 \cite{otov2} and Depgraph \cite{fang2023depgraph}, as well as the classic method Gate Decorator \cite{you2019gate}.
Our main contributions are as follows:

\begin{itemize}[leftmargin=*]
	\item 
	We introduce the \textit{SP} procedure, a multi-step pruning process that gradually decays redundant structures to zero over $N$ steps with ongoing optimization. This approach minimizes drastic network changes and enhances information retention during pruning.

	\item  We develop the  \textit{SR}  procedure, which employs a gradient-driven approach to effectively correct sub-optimal pruning decisions,  enabling more adaptive and optimal pruning decisions.

	\item We demonstrate the effectiveness and generalizability of the proposed DPM by integrating it with three distinct pruning frameworks. Each integration showcases consistent improvements in accuracy and reductions in FLOPs, underscoring DPM's versatility and significant enhancements across various existing pruning methods.

\end{itemize}

\section{Related Work}

\textbf{Soft pruning methods}. 
Several methods are designed to mitigate information loss from pruned structures  by enabling more gentle pruning processes. For instance, SFP \cite{He2018SoftFP} allows the ongoing optimization of pruned weights in subsequent epochs to aid in network recovery. Building upon SFP, ASFP \cite{he2019asymptotic} enhances information retention by gradually raising pruning rate throughout each iteration of pruning. However, the application of a single-step pruning process in these methods — immediately zeroing out pruned filters — results in irreversible information loss and abrupt network changes. Our DPM addresses these drawbacks by progressively reducing the weights of pruning filters to zero, alongside continuous optimization, ensuring improved information preservation and better network adaptation.

\textbf{Dynamic pruning methods.} Dynamic pruning approaches, such as  \cite{lin2020dynamic, evci2021rigging}, dynamically prune and regrow network weights to explore optimal sparsity configurations. These methods are effective at correcting sub-optimal pruning decisions by enabling the regeneration of erroneously pruned weights. However, these methods mainly focus on unstructured pruning and typically employ a single-step pruning process.  Building upon these concepts, our proposed \textit{Self-Rectifying (SR)} procedure leverages gradient information to dynamically identify and correct sub-optimal pruning decisions within a multi-step pruning process. Our approach not only offers a robust, gradient-based criterion for dynamic adjustments but also seamlessly integrates with various existing structured pruning methods to enhance their effectiveness.

\textbf{Auto pruning methods}. Auto pruning frameworks has been popular these year,
due to their advantages of easy deployment and generalizability towards various networks. 
OTOv2 \cite{otov2} is a user-friendly auto pruning framework that prunes while training and without fine-tuning. Given a network, OTOv2 automatically divides it into several zero invariant groups, and prunes these groups with their gradient information while training using a DHSPG strategy. Different from OTOv2, Depgraph  \cite{fang2023depgraph} prunes network through classic iteration of pruning and fine-tuning. This method prunes groups with its own global wise criteria with sparse learning, while also supporting extra criteria including L1, L2,  bn scalar and more. APIB  \cite{10378349} is a novel auto pruning method that prunes with a complex criteria based on Information Bottleneck Principle and achieved state-of-the-art results.

\section{\algname} 
This section introduces the Decay Pruning Method (\algacro{}), which consists of two key components: \textit{Smooth Pruning (SP)} and \textit{Self-Rectifying (SR)}.  Specifically, \textit{SP} replaces traditional single-step pruning, and applies a multi-step pruning process, gradually decaying weights of redundant structures to zero over $N$ steps with ongoing optimization. Meanwhile, \textit{SR} utilizes gradient information from the decaying weights to adaptively identify and correct sub-optimal pruning decisions, considering the ongoing evolution of the network.

\subsection{Smooth Pruning Procedure}
\textit{SP} innovates beyond traditional single-step pruning methods by employing a more gradual $N$-step process. 
Given a pruning structure  characterized by its weight matrix $\bm{x}$, \textit{SP} begins by measuring  its current L2 norm as the initial length, denoted as $L_{init}:=\vert\vert \bm{x} \vert\vert_2$.  Rather than abruptly removing the weights $\bm{x}$, \textit{SP} gradually reduces them to zeros across $N$ iterations, each coinciding with an subsequent optimization step  in the fine-tuning phase. Specifically, at the $k^{th}$ optimization step,  where \(\bm{x}_k\) represents the current state of the weights, \textit{SP} first updates $\bm{x}_k$  to a temporary state as $\widetilde{\bm{x}}_{k+1} := \bm{x}_k-a_k\gg(\bm{x}_k)$ using a  general optimizer, where  $a_k$ is the learning rate and $\gg(\bm{x}_k)$ represents the gradient of $\bm{x}_k$. It then obtains $\bm{x}_{k+1}$ by scaling $\widetilde{\bm{x}}_{k+1}$ to a specified L2 norm, $L_{target}$. This scaling preserves the directional integrity of the weights, ensuring a continuing directional optimization:
\begin{equation}
	\bm{x}_{k+1} \gets \bm{\widetilde{x}}_{k+1} \times \frac{L_{target}}{\vert\vert \bm{\widetilde{x}}_{k+1}\vert\vert_2}
	\label{Xk+1_eqn}
\end{equation}
To ensure a consistent decrement in weight magnitude across iterations,  $L_{target}$ is derived for each iteration based on the progression through the $N$-step process, adhering to the following formula: 
\begin{equation}
	L_{target} := (N-n\_step) \times L_{s},
	\label{Ltarget_eqn}
\end{equation}
where $n\_step$ indicates the current iteration step within the decaying process, systematically increasing from 1 to $N$. This formula emphasizes the sequential reduction in weight magnitude, with $L_s$ dictating the decrement length per iteration. $L_s$ itself is calculated from the initial L2 norm $L_{init}$,  divided by the total decaying steps $N$:
\begin{equation}
	L_{s} :=\frac{ L_{init}}{N}
	\label{Ls_eqn}
\end{equation}
Overall, \textit{SP} ensures an equal reduction in the magnitude of $\bm{x}$  across $N$ iterations. Upon reaching the final step ($n\_step = N$), $\bm{x}$ is zeroed out as $L_{target} = 0$, marking its last update. Otherwise,  \textit{SP} concludes the current iteration by incrementing $n\_step$ by one for next iteration of decaying. Therefore, \textit{SP} achieves a gradual reduction in magnitude of pruning structures while facilitating a ongoing directional optimization, which ultimately enhances the network adaptability towards pruning and better information retention. 




\begin{wrapfigure}{r}{0.5\textwidth}
	\vspace{-7.7mm}
	\centering
	\includegraphics[width=\linewidth]{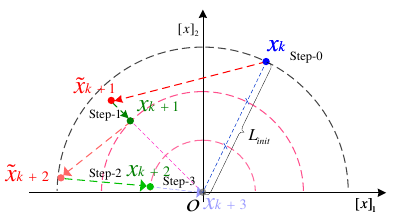}
	\vspace{-5.5mm}
	\caption{ An illustration of the smooth pruning procedure (\textit{SP}) with $N=3$.}
	\label{DPM_pruning}
	\vspace{-3mm}
\end{wrapfigure}

Figure \ref{DPM_pruning} illustrates an example of the \textit{SP} procedure with $N=3$. This figure shows how the weights $\bm{x}_k$  are gradually reduced to zero over three optimization steps. The semi-circular dashed lines indicate the potential ranges where $\bm{x}_k$ can land at each decaying step, constrained by $L_{target}$. 
Initially, at Step-0, \textit{SP}  begins the decaying process by recording the current L2 norm as $L_{init}:=\vert\vert \bm{x}_{k}\vert\vert_2$. In the subsequent optimization steps, \textit{SP} first pre-updates $\bm{x}_k$ to $\widetilde{\bm{x}}_{k+1} \gets \bm{x}_k-a_k\gg(\bm{x}_k)$. Following this,  \textit{SP} scales $\widetilde{\bm{x}}_{k+1}$ to $\bm{x}_{k+1}$ according to Equation \ref{Xk+1_eqn}, aiming for the decayed L2 norm $L_{target}$, as specified by the Equation \ref{Ltarget_eqn}.  This iterative reduction of $\bm{x}_k$ continues until it reaches zero by Step-3, marking its last update.


During the decaying process, a specific condition occurs when the optimized weights $\bm{\widetilde{x}}_{k+1}$ already meet or fall below the intended $L_{target}$, evidenced by  $\vert\vert \bm{\widetilde{x}}_{k+1}\vert\vert_2 \leq L_{target}$.
In such instances, further scaling $\bm{\widetilde{x}}_{k+1}$ to match  $L_{target}$ becomes unnecessary. Therefore, \textit{SP} directly updates $\bm{x}_{k+1}$ as  $\bm{\widetilde{x}}_{k+1}$  without any additional modification. This iteration concludes by recalibrating  the current step ($n\_step$) to align with $\vert\vert \bm{\widetilde{x}}_{k+1}\vert\vert_2 \leq 	L_{target}  \gets  (N-n\_step) \times L_{s} $  for the next iteration, thereby maintaining efficient weight reduction.




\subsection{Self-Rectifying Procedure} 
\textit{SR} leverages gradient information to identify and rectify sub-optimal pruning decisions within the \textit{SP}  procedure. Specifically, at the $k^{th}$ optimization step, the gradient $\gg(\bm{x}_k)$ of the weights $\bm{x}_k$ is decomposed into two components: the direction-wise component $\gg_{d}(\bm{x}_k)$ and the magnitude-wise component $\gg_{m}(\bm{x}_k)$. These components  optimize the direction and magnitude of the weights $\bm{x}_k$, respectively.  Within the  \textit{SP}  procedure, while optimization is allowed directionally, the magnitude of the pruning structures is compulsorily reduced. During this process, if an important structure is incorrectly pruned, or the pruning decisions become less optimal due to network evolution, a strong $\gg_{m}(\bm{x}_k)$ may emerge, attempting to restore the magnitude of the weights. This strong $\gg_{m}(\bm{x}_k)$ indicates resistance against the ongoing decay. Two criteria are proposed to measure $\gg_{m}(\bm{x}_k)$ for timely identification and adjustment of sub-optimal pruning decisions.

\begin{wrapfigure}{r}{0.26\textwidth}
	\vspace{-10mm}
	\centering
	\includegraphics[width=\linewidth]{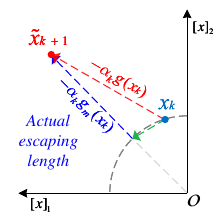}
	\vspace{-5mm}
	\caption{  Illustrations of Actual escaping length.}
	\label{Actual_escape}
	\vspace{-6mm}
\end{wrapfigure}

\textbf{Criterion 1: Actual Escaping Rate. } 
To assess the resistance to pruning, we define the Actual Escaping Length as $\vert\vert \bm{\widetilde{x}}_{k+1} \vert\vert_2 - \vert\vert \bm{x}_{k} \vert\vert_2$. This measure, representing the increase in the L2 norm of the updated weights $\bm{\widetilde{x}}_{k+1}$ driven by $\gg_m(\bm{x}_k)$, also indicates the resistance against the decaying process, as illustrated in Figure \ref{Actual_escape}. 
Given the variability in gradient magnitude across different layers and training phases,  directly employing this measurement as a criterion could result in inconsistencies. To mitigate this, we introduce the Actual Escaping Rate ($C_{rate}$), which normalizes the Actual Escaping Length by the total impact of the gradient on the weights:

\begin{equation}
	C_{rate} := \frac{\vert\vert \bm{\widetilde{x}}_{k+1} \vert\vert_2 - \vert\vert \bm{x}_{k} \vert\vert_2}{\vert\vert \bm{\widetilde{x}}_{k+1} - \bm{x}_{k} \vert\vert_2}
	\label{C_rate_eqn}
\end{equation}

In this equation, the denominator represents the L2 norm of  the total gradient impact, $a_k\gg(\bm{x}_k)$, depicted as the red line in Figure \ref{Actual_escape}. The numerator captures the change in L2 norm of the weights induced by $a_k\gg_{m}(\bm{x}_k)$, depicted as the blue line in Figure \ref{Actual_escape}. This equation ensures that  $C_{rate}$ provides a stable measurement of resistance to decay across various layers and training phases. A higher $C_{rate}$ suggests a stronger resistance to the pruning process, potentially indicating a need to release the weights from further decay.


\textbf{Criterion 2: Magnitude of Gradient. }
While a high $C_{rate}$ can indicate significant resistance to pruning, the magnitude of the gradients should also be considered. However, $C_{rate}$ can not directly identify short gradients.  For instance, a small gradient holding the same direction with the vector pointing from zero to $\bm{x}_{k}$, can still yield a high  $C_{rate}$ of 1, indicating only weak escaping strength and potentially leading to erroneous releases. As such, we designed a dynamic criterion, denoted as  $C_{len}$, which measures the relative magnitude of the gradient compared to its counterparts in parallel channels within the same layer. Specifically, we calculate the mean magnitude of the gradient across parallel channels that share similar structures and weight magnitudes. The $C_{len}$ is defined as follow:

\begin{equation}
	C_{len} := \frac{\vert\vert \gg(\bm{x}_k) \vert\vert_2}{\frac{1}{M} \sum_{j=1}^{M} \vert\vert\gg( \bm{x}_{k}^j) \vert\vert_2}
	\label{C_len_eqn},
\end{equation}

where $M$ represents the number of parallel channels in the same layer, and $\bm{x}^j$ denotes their respective weights. This criterion assesses the  strength of the current gradient relative to the parallel gradients, complementing Criterion 1 by setting a dynamic threshold to ensure substantial escaping strength.


Finally, two hyperparameters, $ T_{rate}$ and  $T_{len}$, are used to determine if the decaying weights $\bm{x}_k$  should be released from the pruning process based on the following condition:

\begin{equation}
	C_{rate} > T_{rate} \quad \land \quad C_{len} > T_{len},\label{met}
\end{equation}

where $\land$ refers to the logical “AND” operation. Once  $\bm{x}_k$ is released from decaying, it returned to normal optimization. The releasing rate can be adjusted by modifying  $T_{rate}$ and $T_{len}$. 

	\textbf{Special Situations.}
	Some pruning methods apply magnitude penalization while fine-tuning. This approach generates an additional force directly targeting zero counteracts most of the escaping strength from the original gradient.
	Consequently, this results in a low $C_{rate}$, complicating the trigger for releasing. In response to this situation, we can either neutralize the effect of penalization when computing $\bm{\widetilde{x}}_{i}$ or consider setting a lower $T_{rate}$.

	%

\subsection{DPM Integration}\label{DPM Integration}
In this section, we discuss  the integration of the Decay Pruning Method into existing pruning frameworks to enhance their efficacy by replacing the conventional single-step pruning process. Given a model $\mathcal{M}$, and its structures $\bm{x}^i \in \mathcal{M}$, general pruning methods usually follow a three-stage iteration process: \textit{(i) {Discovering Pruning Decisions ($\bm{P}$)}:} Decide which structures to prune. \textit{(ii) {Removing Redundant Structures ($\bm{x}^i \in \bm{P}$)}:}  Remove the identified redundant structures.  \textit{(iii) {Fine-Tuning Networks}:}  Adjust the network to recover performance after pruning. Typically, the single-step pruning process operates stage \textit{(i)}  and  \textit{(ii)} concurrently, directly removing identified redundant structures. Differently, DPM modifies this process by interleaving the removal of redundant structures (stage \textit{(ii)}) within the fine-tuning phase (stage \textit{(iii)}). This adjustment allows for a more gradual removal of structures over $N$ steps, facilitating a better identification for redundant structures.  Specifically, DPM consists of an \textit{Initial Phase}, followed by an iterative \textit{Pruning Decision Phase} and a \textit{Decay Pruning Phase} that prunes while fine-tuning.

\begin{minipage}{\linewidth}
	\begin{algorithm}[H]
	\caption{Outline of the \textit{Pruning Decision Phase}.}\label{pruning decision phase}
	\begin{algorithmic}[1]
		\State \textit{\textbf{Pruning Decision Phase:}}
		\State Make pruning decision list $\bm{P}$;
		\For{$\bm{x}^i \in \mathcal{M}$}
		\If{$\bm{x}^i \in \bm{P}$}
		\State \st{Project  $\bm{x}^i$ to zero and never update. }
		\State  \textit{\textbf{Initialize decay process for $\bm{x}^i$ if not already decaying:}}
		\If{\textbf{not} $is\_decay^i$}
		\State $L_{init}^i \gets \|\bm{x}^i\|_2$;
		\State $is\_decay^i \gets True$; \Comment{Set decay\label{trigger_DPM_here}}
		\EndIf
		\EndIf
		\EndFor
	\end{algorithmic}
\end{algorithm}
\end{minipage}

\textbf{Initial Phase.} DPM initializes three variables for each structure $\bm{x}^i$, including: $n\_step^i :=0$, tracking the number of the decaying steps for the structure; $L_{init}^i :=0$, denoting the  initial L2 norm of the structure; and $is\_decay^i:=False$,  indicating whether the structure is currently undergoing decay.

\textbf{Pruning Decision Phase (Algorithm \ref{pruning decision phase}).} During each iteration, DPM calculates the current L2 norm of each redundant structure $\bm{x}^i \in \bm{P}$ , storing it as  $L_{init}^i $, and sets their $is\_decay^i$ as  $True$,  marking the beginning of the decaying process.

\textbf{Decay Pruning Phase (Algorithm \ref{Outline of Decay Pruning Method}).} DPM progressively reduces the magnitude of the redundant structures during optimization steps, in parallel   with fine-tuning. Specifically, at the $k^{th}$ optimization step, for a given decaying structure $\bm{x}_{k}^i$, DPM initially pre-updates it to $\bm{\widetilde{x}}_{k+1}^i$ using standard optimization. After that, two criteria: $C_{rate}$ and $C_{len}$, are calculated based on gradient information to determine whether to continue the decay. If both criteria are met, the decay is halted, and $\bm{x}^i$ is released from the decaying process, returning to the normal optimization process. Otherwise, \textit{SP} projects $\bm{\widetilde{x}}_{k+1}^i$ to achieve a predetermined $L_{target}$, and ultimately updates $\bm{x}_{k+1}^i$ with $\bm{\widetilde{x}}_{k+1}^i$. This iterative process continues until $\bm{x}^i$ fully decays to zero, after which $\bm{x}^i$ is either no longer updated or is directly removed from the model.

\begin{algorithm}[H]
	\caption{Outline of the\textit{ Decay Pruning Phase}.}\label{Outline of Decay Pruning Method}
	\begin{algorithmic}[1]
		\label{Decay Prune with Self-Adapting Procedure}
		\State\textit{ \textbf{In the $k^{th}$ optimization step of fine-tuning: } }
		\For{weights $\bm{x}_k^i \in \mathcal{M}$}
		\State {Pre-update} $\bm{x}_{k}^i $ as $\widetilde{\bm{x}}_{k+1}^i := \bm{x}_{k}^i-a_k\gg(\bm{x}_{k}^i)$.
		\If { $is\_decay^i=True$} 
		\If{$n\_step^i= N$}
		\State $\bm{\widetilde{x}}_{k+1}^i  \gets  0$
		\ElsIf {$n\_step^i < N$ }
		\State \tikzmark{top0}\tikzmark{left0}  Compute $C_{rate}$ and $C_{len}$ using gradients. 
		\Comment{See Eqns. \ref{C_rate_eqn}, \ref{C_len_eqn}}
		\If {$C_{rate} > T_{rate}  \land C_{len} > T_{len}$}  	 	\Comment{{Condition \ref{met}}}	
		
		\State 	\textit{ \textbf{Release $\bm{x}_{k}^i$ from decaying:}}
		\State  $is\_decay^i \gets False$
		\State $n\_step^i  \gets  0$ \tikzmark{bottom0}
		\Else 
		\State \tikzmark{left1}\tikzmark{top1}	$L_{s}  := \frac{L_{init}^i }{N}$ \Comment{\textit{Eqn \ref{Ls_eqn}}}	
		\State   $L_{target}  :=  (N-n\_step^i) \times L_{s} $ 	\Comment{\textit{Eqn \ref{Ltarget_eqn}}}	
		\State Project $\bm{\widetilde{x}}_{k+1}^i  \gets \bm{\widetilde{x}}_{k+1}^i \times \frac{L_{target}}{\vert\vert \bm{\widetilde{x}}_{k+1}^i\vert\vert_2}$	\Comment{\textit{Eqn \ref{Xk+1_eqn}}}	
		
		\State $n\_step^i \gets  n\_step^i+1$ \tikzmark{bottom1}	
		\EndIf
		\EndIf
		\EndIf
		\State Update $\bm{x}_{k+1}^i \gets \bm{\widetilde{x}}_{k+1}^i$.
		\EndFor
	\end{algorithmic}
	\AddNote{top0}{bottom0}{left0}{\vspace{-4.5mm}\\ Self-\\\vspace{-0.8mm}Rectifying}
	\AddNote{top1}{bottom1}{left1}{\vspace{-4.5mm}\\ Smooth\\\vspace{-1mm}Pruning}
\end{algorithm}

\section{Experiment}

In this section, we demonstrate the  effectiveness and generalizability  capabilities of the Decay Pruning Method (DPM) by  integrating it with various pruning frameworks, including the auto-pruning frameworks OTOv2 \cite{otov2} and Depgraph \cite{fang2023depgraph}, as well as the conventional pruning method, Gate Decorator \cite{you2019gate}. We conducted comprehensive experiments across various models such as VGG16  \cite{VGG16}, VGG19, ResNet50 \cite{ResNet50}, and ResNet56 \cite{ResNet50}, utilizing well-known datasets including CIFAR10 \cite{CIFAR10}, CIFAR100 \cite{CIFAR10}, and ImageNet \cite{ImageNet}. These experiments adhered strictly to the original training settings and codes to ensure a valid comparison of DPM's enhancements. 

%
%

For the hyperparameter in \textit{SP}, we set the decaying step  $N$  to 5. For the two criterion hyperparameters in \textit{SR}, \(T_{rate}\) is set between [0.2, 0.65], with different values for different methods and datasets, while \(T_{len}\) is fixed to 0.2 across all approaches and datasets. This ensures that only structures with significant resistance are released from the pruning process. For methods utilizing sparse learning like L1 or L2 penalization, these hyperparameters are set to lower values: 0.1 to 0.2 for \(T_{rate}\) and 0.1 for \(T_{len}\), reducing the impact on the gradient magnitude component $\gg_{m}$. For further details on hyperparameter validation, please refer to Appendix \ref{hyperparameter_verification}.

Our experiments utilized NVIDIA RTX4090 GPUs. In this section, we use '+\textit{SP}' to denote the exclusive use of \textit{SP}, and '+\textit{SR}' when both \textit{SP} and \textit{SR} are applied to each pruning method.  We also benchmarked against various state-of-the-art methods, with results cited from the corresponding literature. Cases with unreported results were noted with ‘-’, and the best pruning results are marked in bold.

\subsection{Integrating DPM with OTOv2} 
OTOv2 \cite{otov2} is a popular auto-pruning method that prunes network while training, without any additional fine-tuning. This method applies a single-step pruning process, directly projecting the redundant structures to zeros and never update. We enhanced OTOv2 by replacing its single-step pruning process with \textit{SP}, and introduce \textit{SR}  to distinguish and rectify sub-optimal pruning decisions utilizing gradient information. To validate the effectiveness of these innovations, we replicated key experiments from OTOv2, including trials with VGG16 on CIFAR10, VGG16-BN on CIFAR10, ResNet50 on CIFAR10, and ResNet50 on ImageNet. These experiments were conducted using the original training codes and maintained consistent hyperparameter settings.

\begin{table}[h]
	\vspace{-1mm}
	\centering
	\caption{VGG16 and VGG16-BN on CIFAR10.} 

\label{table.vgg_cifar}

\resizebox{\textwidth}{!}{
	\begin{tabular}{c|c|c|c|c|c|c|c|c|c}
		\Xhline{3\arrayrulewidth}
		Method & BN & FLOPs & Params &  Top-1 Acc.&Method & BN & FLOPs & Params &  Top-1 Acc. \\
		\hline
		Baseline & \xmark  & 100\% & 100\% & 91.6\%& Baseline & \cmark &  100\% & 100\% & 93.2\%  \\
		SBP \cite{Neklyudov2017StructuredBP} &  \xmark& 31.1\% & 5.9\% &  91.0\% & EC \cite{li2017pruning} &  \cmark &  65.8\% & 37.0\% & 93.1\%  \\
		BC \cite{Louizos2017BayesianCF}&  \xmark & 38.5\% & 5.4\% &  91.0\%  & Hinge \cite{Li2020GroupST} & \cmark & 60.9\% & 20.0\% & 93.6\%   \\
		RBC \cite{Zhou2018AccelerateCV} & \xmark &  32.3\% & 3.9\% & 90.5\% &SCP \cite{Kang2020OperationAwareSC} & \cmark & 33.8\% & 7.0\% & \textbf{93.8\%}  \\
		RBP \cite{Zhou2018AccelerateCV} & \xmark &  28.6\% & 2.6\% &  91.0\%&OTOv1 \cite{otov1} & \cmark & 26.8\%  & 5.5\%  & 93.3\%  \\
		OTOv1 \cite{otov1} & \xmark &  {16.3\%} & 2.5\% & 91.0\%&OTOv3 \cite{otov3} & \cmark & 26.6\%  & 5.0\%  & 93.4\%   \\
		
		\hdashline

		OTOv2 \cite{otov2}  & \xmark &  \textbf{13.0\%} &
		\textbf{2.3\%} & 91.65\%& OTOv2 \cite{otov2}  & \cmark & 26.5\% & \textbf{4.8\%} & 93.4\%\\ 
		+\textit{SP} (Ours)    & \xmark & 13.4\% & 2.5\% & 92.11\%&  +\textit{SP} (Ours)   & \cmark & 26.4\% & \textbf{4.8}\% & 93.6\%  \\
		+\textit{SR}  (Ours) & \xmark & 15.0\% & 2.8\% & \textbf{92.65\%} & +\textit{SR} (Ours)  & \cmark & \textbf{25.8}\% & \textbf{4.8}\% & \textbf{93.8\%}   \\
		
		\Xhline{3\arrayrulewidth} 
	\end{tabular}
	
}

\end{table}

\textbf{VGG16 for CIFAR10.} We evaluated \algacro{} on CIFAR10 using both the vanilla VGG16 and a variant known as VGG16-BN, which includes a batch normalization layer after each convolutional layer. The results, presented in Table \ref{table.vgg_cifar}, contrast the original OTOv2 with its enhanced version incorporating DPM. Specifically, for the vanilla VGG16, the \textit{SP} component of \algacro{} yielded a 0.4\% boost in accuracy, albeit with a slight increase in FLOPs and parameters. By further incorporating the \textit{SR} component, \algacro{} achieved a noteworthy 1\% improvement in accuracy,  resulting in a superior accuracy-efficiency trade-off. For the VGG16-BN variant, \algacro{} achieved a top-1 accuracy of 93.8\%, utilizing even fewer FLOPs, clearly surpassing the performance of the original OTOv2 and other competitive methods such as OTOv3 and SCP.

\begin{wraptable}{r}{0.55\textwidth}	
\vspace{-7mm}
\begin{minipage}{\linewidth}
	\caption{ResNet50 on CIFAR10}
	\centering
	\resizebox{\textwidth}{!}{
		\begin{tabular}{ c|c|c|c}
			\Xhline{3\arrayrulewidth}
			Method	&	FLOPs	& Params	&	Top-1 Acc.\\
			\hline
			Baseline				&  100\%       &	100\%      	  &	93.5\%    \\
			AMC \cite{He2018AMCAF}  & –			   & 60.0\% 		  &	93.6\%    \\
			PruneTrain \cite{Lym2019PruneTrainFN} &	30.0\%	&–	&93.1\% \\
			N2NSkip \cite{Subramaniam2022N2NSkipLH} &	– &	10.0\%&	94.4\% \\
			OTOv1 \cite{otov1} &	12.8\%&	8.8\%&	94.4\% \\
			\hline
			OTOv2 \cite{otov2} (90\% group sparsity) & 2.2\% & 1.2\% & 93.0\% \\
			+\textit{SP}  (90\% group sparsity)& 2.16\% & 0.84\% & 92.9\% \\
			+\textit{SR} (90\% group sparsity)&  \textbf{1.7\%} &  \textbf{0.8\%} & 93.0\%\\	
			\hline
			OTOv2 \cite{otov2} (80\% group sparsity)& 7.8\% & 4.1\% & 94.5\% \\
			+\textit{SP}  (80\% group sparsity)& 7.8\% & 3.3\% & 94.66\% \\
			+\textit{SR} (80\% group sparsity)&  6.9\% &  3.1\% &  \textbf{94.70\%} \\
			\Xhline{3\arrayrulewidth}
	\end{tabular}}
	\label{table.resnet50_cifar10}
\end{minipage}
\vspace{-3mm}
\end{wraptable}

\textbf{ResNet50 for CIFAR10.} We evaluated our DPM using ResNet50 on CIFAR10, as detailed in Table ~\ref{table.resnet50_cifar10}. 
While the OTOv2 method already provides state-of-the-art results with extreme pruning rates, \algacro{} has further enhanced this by achieving a better balance between accuracy and model efficiency. Specifically, under 90\% group sparsity, DPM  impressively reduced the FLOPs to 1.7\% and parameters to 0.8\%, without sacrificing performance. Under 80\% group sparsity,  \algacro{} further decreased FLOPs to 6.9\% and parameters to 3.1\%, while still achieving a top-1 accuracy of 94.7\%. These results are highly competitive to other pruning methods such as OTOv1 and N2NSkip, yet achieves 2-3 times more efficiency. 

\begin{wraptable}{r}{0.55\textwidth}	
\centering
\vspace{-5mm}
\begin{minipage}{\linewidth}
	\caption{ResNet50 on ImageNet}
	\centering
	\resizebox{\textwidth}{!}{
		\begin{tabular}{c|c|c|c}
			\Xhline{3\arrayrulewidth}
			Method & FLOPs & Params & Top-1/5 Acc. \\
			\hline
			Baseline & 100\% & 100\% & 76.1\% / 92.9\% \\
			CP \cite{He2017ChannelPF}& 66.7\% & -- & 72.3\% / 90.8\% \\
			DDS-26 \cite{Huang2018AdvancesIT} & 57.0\% & 61.2\% & 71.8\% / 91.9\% \\
			SFP \cite{He2018SoftFP} & 41.8\% & -- & 74.6\% / 92.1\% \\ 
			RRBP \cite{Zhou2018AccelerateCV}  & 45.4\% & -- & 73.0\% / 91.0\% \\
			Group-HS \cite{Yang2019DeepHoyerLS}  & 52.9\% & -- & \textbf{76.4\%} / \textbf{93.1\%} \\
			Hinge \cite{Li2020GroupST}& 46.6\% & -- & 74.7\% / \ \ -- \ \ \ \ \\ 
			SCP \cite{Kang2020OperationAwareSC} &  45.7\%	& --	& 74.2\% / 92.0\% \\
			ResRep \cite{Ding2020ResRepLC} & 45.5\% & -- & {76.2\%} / 92.9\% \\
			OTOv1 \cite{otov1}& 34.5\% & 35.5\% & 74.7\% / 92.1\% \\
			\hdashline
			OTOv2 \cite{otov2} (70\% group sparsity) & 14.5\% & \textbf{21.3\%} & 70.1\% / 89.3\% \\ 
			+\textit{SP} (70\% group sparsity)& 13.8\% & 22.0\% & 70.7\% / 89.6\%  \\
			+\textit{SR} (70\% group sparsity)& \textbf{13.28}\% & 22.36\% & 70.94\% / 89.63\% \\
			\hdashline
			OTOv2 \cite{otov2} (60\% group sparsity) & 20.0\% & 28.5\% & 72.2\% / 90.7\% \\ 
			+\textit{SP} (60\% group sparsity)& 20.2\% & 29.7\% & 72.75\% / 90.9\% \\
			+\textit{SR} (60\% group sparsity)& 19.82\% & 30.1\% & 72.86\% / 90.8\%   \\
			\Xhline{3\arrayrulewidth}
	\end{tabular} }
	\label{fig:resnet50_imagenet}
\end{minipage}
\vspace{-7mm}
\end{wraptable}

\textbf{ResNet50 for ImageNet.} We assessed the performance of our DPM on the ResNet50 model using the challenging ImageNet dataset. 
In particular, under 70\% group sparsity, DPM achieved a significant 0.84\% increase in top-1 accuracy and a 1.32\% further reduction in FLOPs compared to the original OTOv2 method.  Similarly, under 60\% group sparsity,  \algacro{} yielded a 0.66\% improvement in top-1 accuracy. 

Overall, the integration of DPM within the OTOv2 framework  underscores its substantial effectiveness in enhancing pruning while training methods. DPM consistently delivers significant improvements by either boosting accuracy or further reducing FLOPs across various benchmarks and pruning rates, setting new state-of-the-art results.

\vspace{3mm}

\subsection{Integrating DPM with Depgraph} 
Depgraph \cite{fang2023depgraph} is an automated pruning framework that implements a classic iterative pruning strategy, alternating between pruning and fine-tuning. Given a pruning structure, Depgraph directly removes it from the network, which can result in abrupt network changes and considerable information loss. We address these drawbacks by introducing DPM, which provides a gradually removal of structures and a self-rectifying procedure to adjust pruning decisions adaptively. We integrated DPM with Depgraph following the methodologies outlined in Section \ref{DPM Integration}, where structures are only removed once they have fully decayed to zero. Depgraph supports various established pruning criteria,  including their own proposed magnitude-based global pruning criteria (Group pruner), both with and without sparse learning, and a group-level batch normalization scalar  (BN pruner) adapted from \cite{liu2017learning}. We have validated the effectiveness of DPM across these criteria using benchmarks such as ResNet56 on CIFAR10 and VGG19 on CIFAR100. For clarity in our results, we denote the use of the Group pruner without sparse learning as 'w/o SL'.


\begin{wraptable}{r}{0.5\textwidth}	
\vspace{-5mm}
\centering
\begin{minipage}{\linewidth}
	
	\caption{ResNet56 on CIFAR10}
	\label{table.Depgraph_resnet56_cifar}
	\centering
	\resizebox{\textwidth}{!}{
		\begin{tabular}{ c|c|c|c}
			\Xhline{3\arrayrulewidth}
			
			Method	&	FLOPs	&	Params	& Top-1 	Acc.\\ 
			\hline
			Polar \cite{Tao2020NeuronlevelSP}&  53.2\%&  --& 93.83\% \\ 
			
			SCP \cite{Kang2020OperationAwareSC}&  51.5\%&  48.47\% & 93.23\% \\ 
			Hinge \cite{Li2020GroupST}&  50.0\%&  48.73\% & 93.69\% \\ 
			
			ResRep \cite{Ding2020ResRepLC} & 47.2\% & -- & 93.71\% \\ 
			APIB \cite{10378349}&  46.0\%&  50.0\%& 93.92\% \\ 
			SFP \cite{He2018SoftFP} &  47.4\%&  --& 93.66\% \\ 
			ASFP \cite{he2019asymptotic} &  47.4\%&  --& 93.32\% \\ 
			\hline
			Group pruner& 46.86\% &52.9\% & 93.84\% \\ 
			+\textit{SP} (Ours)  & 46.32\% & 49.69\% & 93.96\% \\ 
			+\textit{SR} (Ours)&  \bf 45.80\%& \bf  47.22\%  & \bf 94.13\%\\ 
			\hline
			Group pruner w/o SL& 47.2\% &69.7\% & 93.32\% \\ 
			+\textit{SP} (Ours)  & 46.7\% & 68.1\% & 93.62\%\\ 
			+\textit{SR} (Ours)&  47.1\%&  65.73\% & 93.71\% \\ 
			\hline
			
			BN  pruner& 46.4\% &58.3\% & 93.50\%\\
			+\textit{SP} (Ours) & 46.8\% & 57.3\% & 93.71\%\\ 
			+\textit{SR} (Ours)&  46.9\%&  56.5\% & 93.77\% \\ 
%
			\Xhline{3\arrayrulewidth}
	\end{tabular}}
\end{minipage}
\vspace{-2mm}
\end{wraptable}

\textbf{ResNet56 for CIFAR10.} We evaluated \algacro{}-enhanced Depgraph using ResNet56 on the CIFAR10 dataset, employing various criteria. As detailed in Table  ~\ref{table.Depgraph_resnet56_cifar}, when utilizing Group pruner with and without sparse learning, DPM significantly boosted accuracy in both scenarios while further reducing parameters by nearly 4\%. Notably, for the Group pruner with sparse learning,  DPM achieved a new state-of-the-art accuracy of 94.13\% while further reducing FLOPs by 1\% and parameters by 5.7\%. This result clearly surpassed the state-of-the-art method APIB by 0.2\% in accuracy, while achieving higher model efficiency. Applying DPM with the BN pruner, a conventional pruning criterion, we observed an additional accuracy improvement of 0.27\%. These results underscore the robustness and adaptability of DPM across different pruning criteria and conditions. Moreover, compared to soft pruning methods such as SFP and its enhanced version ASFP, DPM shows a significant surpass in accuracy and FLOP reduction.


\begin{wraptable}{r}{0.5\textwidth}	
\begin{minipage}{\linewidth}
	\centering
	\vspace{-7mm}
	\caption{VGG19 on CIFAR100}
	\label{table.Depgraph_VGG19_cifar}
	
	\centering
	
	\resizebox{\textwidth}{!}{
		\begin{tabular}{ c|c|c|c}
			\Xhline{3\arrayrulewidth}
			Method	&	FLOPs	&	Params	&	Top-1  Acc.\\ 
			\hline
			EigenD \cite{Wang2019EigenDamageSP} &  11.37\%&-- & 65.18\% \\ 
			
			GReg-1 \cite{Wang2020NeuralPV}&  11.31\%&  -- & 67.55\% \\ 
			GReg-2 \cite{Wang2020NeuralPV}&  11.31\%&  -- & 67.75\% \\ 
			\hline
			Group pruner& 11.03\% &6.36\% & 70.74\% \\
			+\textit{SP} (Ours)  & 10.95\% & \textbf{5.82\%} & 70.85\% \\ 
			+\textit{SR} (Ours)&  \textbf{10.86\%} &  6.10\% & \textbf{70.92\%}\\ 
			\hline
			Group pruner w/o SL& 11.36\% &8.10\% & 67.58\% \\ 
			+\textit{SP}  (Ours) & 12.06\% & 7.82\% & 67.66\% \\ 
			+\textit{SR} (Ours)&  12.06\%&  8.36\% & 67.76\%\\ 
			\Xhline{3\arrayrulewidth}
			
	\end{tabular}}
\end{minipage}
\vspace{-5mm}
\end{wraptable}

\textbf{VGG19 for CIFAR100.} We also validated DPM-enhanced Depgraph with VGG19 on the CIFAR100 dataset, a more challenging benchmark. As shown in Table \ref{table.gate-decorator_resbet50_cifar}, DPM ultimately provided an accuracy enhancement by 0.18\% to the Group pruner, both with and without sparse learning, surpassing other methods in both accuracy and FLOP reduction.

The application of DPM to Depgraph has confirmed the generalizability of \textit{SP} and \textit{SR} across various criteria and conditions, each demonstrating significant improvements from individual perspectives.

\subsection{Integrating DPM with Gate Decorator}

\begin{wraptable}{r}{0.5\textwidth}	
	\vspace{-4mm}
	\begin{minipage}{\linewidth}
		\centering
		\caption{ResNet50 on CIFAR10}
		
		\label{table.gate-decorator_resbet50_cifar}
		\centering
		\resizebox{\textwidth}{!}{
			\begin{tabular}{ c|c|c|c|c}
				\Xhline{2\arrayrulewidth}
				Architecture & Method	&	FLOPs	&	Params	& Top-1 	Acc.\\ 
				\hline
				\multirow{3}{*}{\shortstack{ResNet56}}
				&Original & 29.81\% &\textbf{31.72\%} & 92.63\% \\ 
				&+\textit{SP}  (Ours) & 29.96\% & 32.11\% &92.70\% \\ 
				&+\textit{SR} (Ours)& \textbf{ 29.80\%}&  32.55\% & \textbf{92.82\%}\\ 
				\hline
				\multirow{3}{*}{\shortstack{VGG16}}
				&Original & 9.86\% &1.98\% & 91.50\% \\ 
				&+\textit{SP}   (Ours)& 9.88\% &1.97\% &91.58\% \\ 
				&+\textit{SR} (Ours)&  \textbf{9.79\%}& \textbf{1.95\%} &\textbf{ 91.74\%}\\ 
				\Xhline{2\arrayrulewidth}
		\end{tabular}}
	\end{minipage}
	\vspace{-2mm}
\end{wraptable}

Gate Decorator \cite{you2019gate} is a classic mask-based pruning method that  leverages the magnitude of batch normalization (BN) scalars to guide pruning decisions. Unlike the previously discussed auto-pruning frameworks, Gate Decorator groups structures by their BN layers and prunes by zeroing out their BN masks. In our integration with DPM, we decay and assess the gradients from the entire weights of each structure, not just the BN masks. This broader modification provides more robust gradient information for \textit{SR} to accurately identify and release sub-optimal pruning decisions.

We evaluated the DPM-enhanced Gate Decorator on VGG16 and ResNet56 using the CIFAR10 dataset. This  integration of DPM resulted in an accuracy increase of 0.19\% for ResNet56 and 0.24\% for VGG16. Specifically, \textit{SP}  contributed to a more gradual pruning process, which improved accuracy by approximately 0.07\%. \textit{SR} further enhanced this by correcting sub-optimal decisions. Notably, for VGG16, \textit{SR}  achieved an additional accuracy improvement of 0.16\% with higher model efficiency over \textit{SP} by releasing only three sub-optimal pruning decisions.   These results underscore the broad applicability and effectiveness of DPM in enhancing mask-based pruning methods.




\section{Conclusion}
This paper introduced the Decay Pruning Method (DPM), a novel smooth pruning approach  that prunes by gradually decaying redundant structures over multiple steps while rectifying sub-optimal pruning decisions using gradient information.  Tested across diverse pruning frameworks like OTOv2, Depgraph, and Gate Decorator, DPM has demonstrated exceptional adaptability and effectiveness, improving performance across various pruning strategies and conditions.  These results highlights DPM's potential as a versatile tool that can be embedded on various existing pruning methods and provides significant enhancements.

%
%

\bibliographystyle{plain}
\bibliography{reference.bib}

\appendix

\section{Hyperparameter Verification}\label{verify_N}
\label{hyperparameter_verification}
In this appendix, we explore the effects of hyperparameters of DPM, including the decaying step $N$ for \textit{SP}, and two thresholds $T_{rate}$ and $T_{len}$  for \textit{SR}, verifying them on benchmarks with OTOv2 and Depgraph.

	\begin{figure}[h]
		\centering
		\begin{subfigure}[b]{0.35\linewidth}
			\centering
			\includegraphics[width=0.9\linewidth]{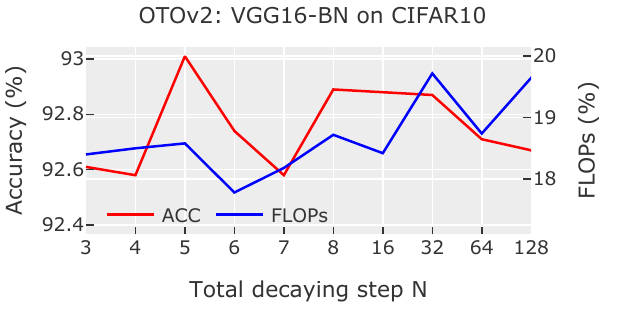}
			\caption{OTOv2: VGG16-BN on CIFAR10}
			\label{OTOv2: VGG16-bn for CIFAR10}
		\end{subfigure}
		\begin{subfigure}[b]{0.35\linewidth}
			\centering
			
			\includegraphics[width=0.9\linewidth]{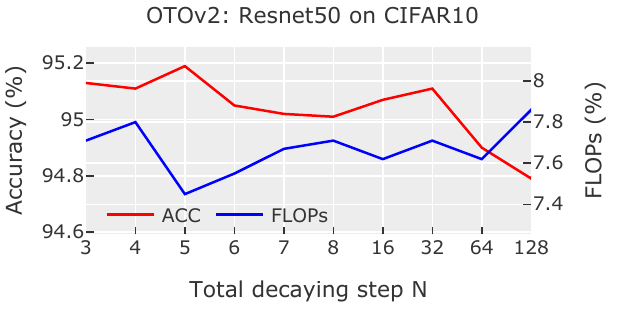}
			\caption{OTOv2: ResNet50 on CIFAR10}
			\label{OTOv2: Resnet50 for CIFAR10}
		\end{subfigure}
		\begin{subfigure}[b]{0.35\linewidth} 
			\centering
			\includegraphics[width=0.9\linewidth]{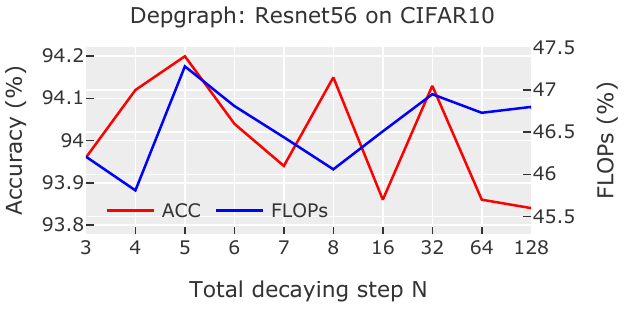}
			\caption{Depgraph: ResNet56 on CIFAR10}\label{Depgraph: Resnet56 for CIFAR10}
		\end{subfigure}
		\begin{subfigure}[b]{0.35\linewidth}
			\centering
			
			\includegraphics[width=0.9\linewidth]{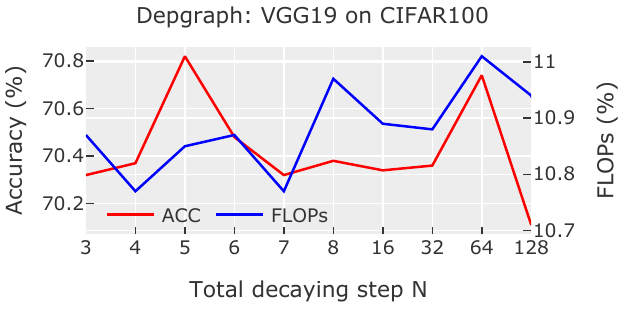}
			\caption{Depgraph: VGG19 on CIFAR100}
			\label{Depgraph: VGG19 for CIFAR100}
		\end{subfigure}
		\caption{Illustrations of the impact of varying decaying step $N$ from 3 to 128. Red lines indicate the accuracy, and blue lines represent the FLOPs of the pruned networks. }	\label{N}
	\end{figure}

1) \textbf{Hyperparameter $N$}: This hyperparameter dictates the number of the optimization steps that \textit{SP} utilizes to decay a pruning structure to zero. We vary $N$ from 3 to 128 to encompass both small and large value conditions. The experiments are conducted on four different benchmarks for OTOv2 and Depgraph, focusing on only \textit{SP}-enhanced conditions. The results are depicted in Fig. \ref{N}. We find that  setting \(N\) too high   leads to diminished accuracy and increased computational and time overheads. Thus, smaller $N$ is encouraged. Overall, setting $N$ to 5 brings in the highest accuracy and relatively low FLOPs from Fig. \ref{N}(a)-(d), providing a better trade-off between pruning performance and efficiency.

		\begin{figure}[h]
		\centering
		\begin{subfigure}[b]{0.35\linewidth}
			\centering
			
			\includegraphics[width=0.9\linewidth]{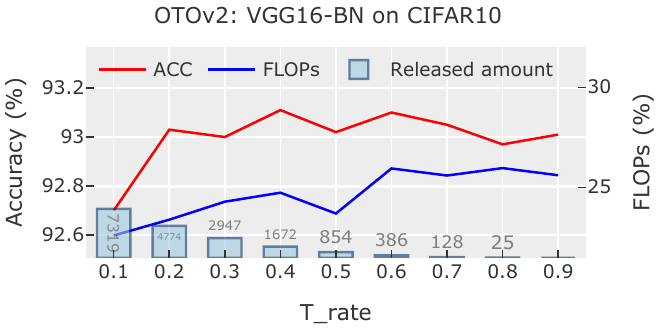}
			\caption{OTOv2: VGG16-BN for CIFAR10}
			\label{T_len_for_VGG16_CIFAR10_OTOv2}
			
		\end{subfigure}
		\begin{subfigure}[b]{0.35\linewidth}
			\centering
			
			\includegraphics[width=0.9\linewidth]{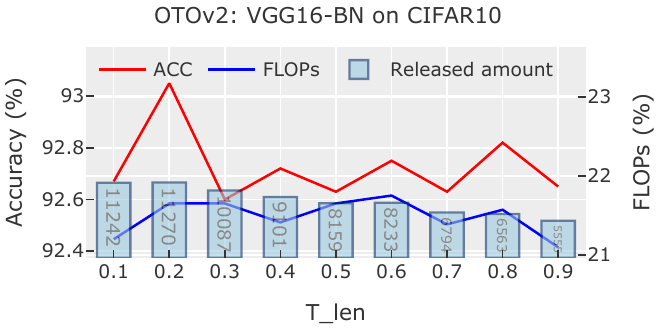}
			\caption{OTOv2: VGG16-BN for CIFAR10}\label{T_rate_for_VGG16_CIFAR10_OTOv2}
		\end{subfigure}
		\begin{subfigure}[b]{0.35\linewidth} 
			\centering
			\includegraphics[width=0.9\linewidth]{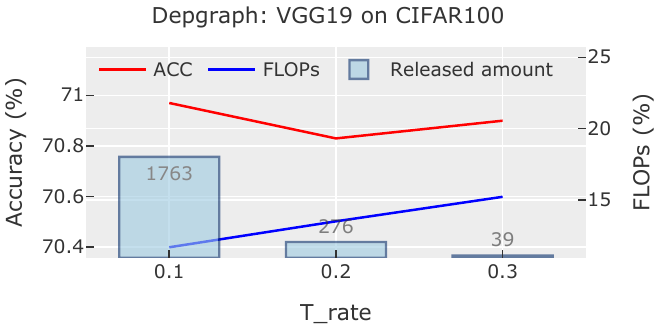}
			\caption{Depgraph: VGG19 for CIFAR100}\label{T_len_for_VGG19_CIFAR100_Depgraph}
		\end{subfigure}
		\begin{subfigure}[b]{0.35\linewidth}
			\centering
			
			\includegraphics[width=0.9\linewidth]{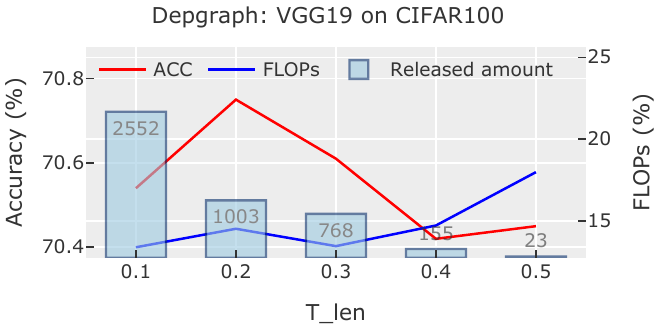}
			\caption{Depgraph: VGG19 for CIFAR100}\label{T_rate_for_VGG19_CIFAR100_Depgraph}
		\end{subfigure}
		\caption{Illustrations of the impact of varying $T_{rate}$ and $T_{len}$. Red lines indicate the accuracy, and blue lines represent the FLOPs of the pruned networks. The total released amount is depicted with histograms. }\label{verify_T_len_and_rate}
	\end{figure}
	
2) \textbf{Hyperparameters $T_{rate}$ and $T_{len}$}: These two hyperparameters govern the release amount for sub-optimal pruning structures in the \textit{SR} procedure. We varied both hyperparameters from 0.1 to 1, with increments of 0.1. Results are presented only for cases where the released amount exceeded 0, indicating that \textit{SR} effectively influenced the outcomes. To assess the individual impact of each hyperparameter, we conducted evaluations by setting one parameter to zero while adjusting the other, with the decaying step fixed at $N=5$. This setting allows us to isolate and observe the effects of $T_{rate}$ and $T_{len}$ on accuracy, FLOPs,  and released amounts. The hyperparameters  were evaluated with OTOv2 and Depgraph on two well-established benchmarks: VGG16 on CIFAR10 and VGG19 on CIFAR100. The experimental results are shown in Fig. \ref{verify_T_len_and_rate}.

By comparing Fig. \ref{T_len_for_VGG16_CIFAR10_OTOv2} and \ref{T_len_for_VGG19_CIFAR100_Depgraph} with Fig. \ref{T_rate_for_VGG16_CIFAR10_OTOv2} and \ref{T_rate_for_VGG19_CIFAR100_Depgraph}, we observe that $T_{rate}$ provides a more significant and stable improvement in accuracy compared to $T_{len}$. This indicates that the proposed Actual Escaping Rate offers a more effective criterion for optimal pruning decisions than methods relying solely on gradient magnitude. Moreover, as shown in \ref{T_rate_for_VGG16_CIFAR10_OTOv2} and \ref{T_rate_for_VGG19_CIFAR100_Depgraph}, the  proposed dynamic gradient magnitude, $T_{len}$, results in noticeable accuracy gains especially at a small setting of 0.2. This setting ensures that the gradient magnitude remains sufficient to complement $T_{rate}$, leading to enhanced and stable improvements in accuracy. Based on these findings, we recommend setting $T_{{len}}$ initially to a smaller value, such as 0.1 or 0.2, and adjusting $T_{rate}$ to effectively control the release amount during the pruning process.

%

	\section{Pruning Efficiency Analysis of DPM}\label{efficiency}
	This section demonstrates the pruning efficiency of the \textit{SP} and \textit{SR} procedure of the Decay Pruning Method (DPM). We assess these components individually by tracking the progression of network sparsity and the amount of structures released by \textit{SR} during the pruning process. Our analysis encompasses tests on four benchmarks using OTOv2 and Depgraph, with each configured to $N=5$ and low values settings for $T_{rate}$ and $T_{len}$ to trigger numerous releases.
	
	\begin{figure}[h]
		\centering
		\begin{subfigure}[b]{0.4\linewidth}
			\centering
			\includegraphics[width=0.9\linewidth]{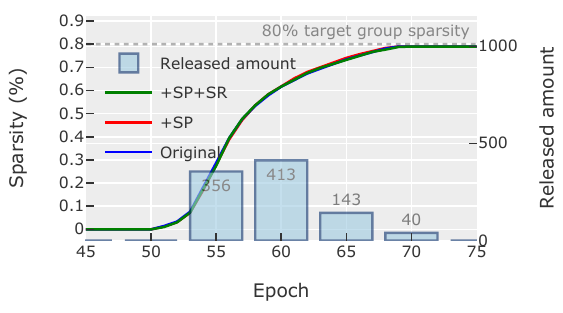}
			\vspace{-2mm}
			\caption{OTOv2: VGG16-BN on CIFAR10}\label{figure.sparsity}
		\end{subfigure}
		\hspace{2mm}
		\begin{subfigure}[b]{0.4\linewidth}
			\centering
			
			\includegraphics[width=0.9\linewidth]{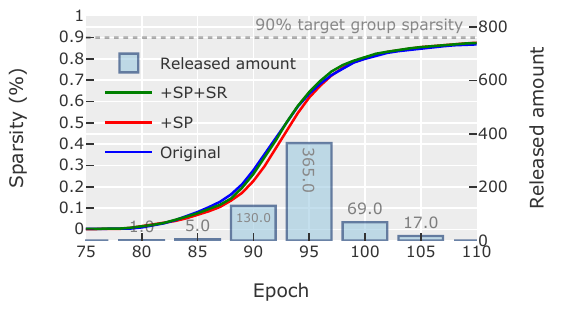}
			\vspace{-2mm}
			\caption{OTOv2: ResNet50 on CIFAR10 }\label{figure.sparsity}
			\label{with penalization}
		\end{subfigure}
		
		\begin{subfigure}[b]{0.4\linewidth}
			\centering
			\includegraphics[width=0.9\linewidth]{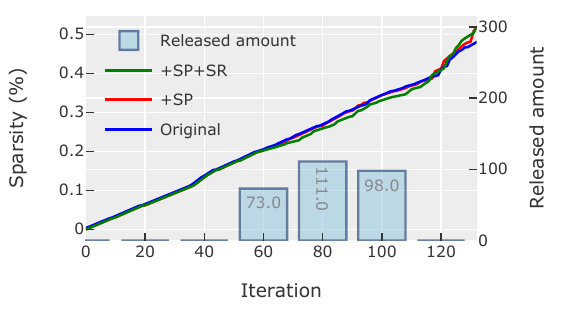}
			\vspace{-2mm}
			\caption{Depgraph: ResNet56 on CIFAR10}\label{figure.sparsity}
			\label{with penalization}
		\end{subfigure}
		\hspace{2mm}
		\begin{subfigure}[b]{0.4\linewidth}
			\centering
			\includegraphics[width=0.9\linewidth]{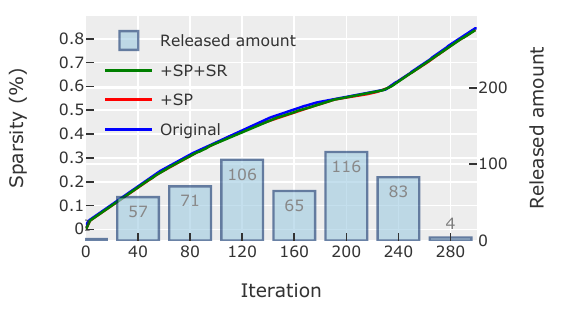}
			\vspace{-2mm}
			\caption{Depgraph: VGG19 on CIFAR100}
		\end{subfigure}
		\caption{Illustration of group sparsity growth during pruning with \textit{SP} and \textit{SR} compared to the original methods. Blue lines represent the original method, while red and green lines indicate the implementations of \textit{SP} and \textit{SP} with \textit{SR}, respectively. The released amount of \textit{SR} accumulated over a certain period is depicted with histograms.}\label{fig:sparsity_growth}
	\end{figure}
	
	As illustrated in Figure \ref{fig:sparsity_growth}, with five steps of decaying, \textit{SP} achieved target sparsity nearly concurrently with the original method, without imposing significant overheads during fine-tuning. Additionally, \textit{SR} effectively rectified hundreds of sub-optimal pruning decisions while still maintaining a high pruning efficiency, with slight or even reduced overheads compared to the original method. Notably, since DPM leverages the fine-tuning phase inherent in the original pruning strategies (e.g., the training phase of OTOv2, which also serves as a fine-tuning phase for pruned structures), it does not necessitate an additional fine-tuning phase. This integration ensures that DPM not only enhances accuracy significantly but also preserves the efficiency of the pruning process.

\end{document}